\documentclass[11pt]{article}
\usepackage[final]{acl}
\usepackage{times}
\usepackage{latexsym}

\usepackage{graphicx}
\usepackage{multirow}
\usepackage{verbatim}
\usepackage{lipsum}
\usepackage{amssymb}
\usepackage{amsmath}
\usepackage{booktabs}
\usepackage{url} 
\usepackage{enumitem}
\usepackage{mathtools}
\usepackage{subcaption}
\usepackage{fontawesome} 
\usepackage{contour}
\usepackage{textcomp}
\usepackage{amsmath}
\usepackage{microtype}

\usepackage{microtype}

\title{Think Before You Speak: Explicitly Generating Implicit Commonsense Knowledge for Response Generation}

\author{
Pei Zhou$^{1}$\thanks{~ Work done while Pei Zhou was an intern at Amazon Alexa AI} \quad Karthik Gopalakrishnan$^{2}$ \quad Behnam Hedayatnia$^{2}$ \quad Seokhwan Kim$^{2}$ \\ \textbf{Jay Pujara$^{1}$ \quad Xiang Ren$^{1}$ \quad  Yang Liu$^{2}$ \quad Dilek Hakkani-Tur$^{2}$}\\
$^1$ Department of Computer Science, University of Southern California\\
$^2$ Amazon Alexa AI\\
\small{\texttt{\{peiz,jpujara,xiangren\}@usc.edu},} \\
\small{\texttt{\{karthgop,behnam,seokhwk,yangliud,hakkanit\}@amazon.com}}
}

\date{}

\begin{document}
\maketitle

\begin{abstract}

Implicit knowledge, such as common sense, is key to fluid human conversations. Current neural response generation (RG) models are trained to generate responses directly, omitting unstated implicit knowledge. In this paper, we present Think-Before-Speaking (TBS), a generative approach to first externalize implicit commonsense knowledge (\emph{think}) and use this knowledge to generate responses (\emph{speak}). 
We expect that externalizing implicit knowledge allows more efficient learning, produces more informative responses, and enables more explainable models. 
We analyze different choices to collect knowledge-aligned dialogues, represent implicit knowledge, and transition between knowledge and dialogues.
Empirical results show TBS models outperform end-to-end and knowledge-augmented RG baselines on most automatic metrics and generate more informative, specific, and commonsense-following responses, as evaluated by human annotators.
TBS also generates \emph{knowledge} that makes sense and is relevant to the dialogue around 85\% of the time.



\end{abstract}

\section{Introduction}\label{intro}

\begin{figure}[tb]
	\centering
	\includegraphics[width=0.95\columnwidth]{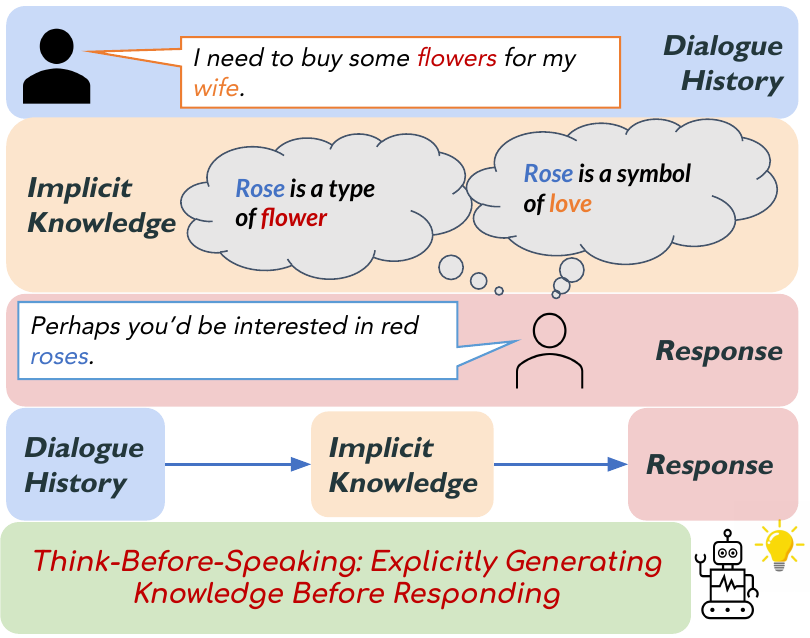}
	\caption{
	{\small \textbf{A motivating example for our study.} We look to train models to externalize the implicit knowledge grounding step by explicitly generating knowledge before responding.}}
	\label{fig:motivation}
\end{figure}

Human communication strives to achieve \textit{common ground}, consisting of mutual beliefs and common knowledge~\cite{stalnaker1978assertion, clark1989contributing}.
Such common ground depends not only on utterances, but also implicit knowledge.
For example, in Figure~\ref{fig:motivation}, this common ground includes the relevant implicit background knowledge ``\emph{rose is a type of flower}''.
Integrating such common ground in utterances is an implicit process often referred to as \emph{knowledge grounding}~\cite{clark1991grounding}.
Recent state-of-the-art neural response generation (RG) models based on pre-trained language models (LM) mostly produce responses in an \emph{end-to-end} manner~\cite{vaswani2017attention, zhang-etal-2020-dialogpt, lewis-etal-2020-bart}, \emph{i.e.}, models are trained to take history and produce a response.
Since implicit knowledge is unstated in dialogue history, RG models do not explicitly learn knowledge grounding  and  may generate uninformative and hallucinated responses~\cite{serban2017hierarchical,welleck2019neural, roller2020recipes}. 
Knowledge-grounded RG~\cite{ghazvininejad2018knowledge,dinan2018wizard, gopalakrishnan2019topical} addresses this issue, however, most approaches require a knowledge base (KB) to \emph{retrieve} knowledge for RG~\cite{zhou2018commonsense,zhao-etal-2020-knowledge-grounded,eric2021multi}, which may suffer from the limited knowledge coverage of the used KBs. 
Some work also casts knowledge as a \emph{latent} factor in generation~\cite{tuan2020knowledge,xu2021retrieval}, which makes it hard to examine the quality of knowledge generation and how \emph{exactly} RG uses the implicit knowledge, posing interpretability concerns.


We propose \emph{Think-Before-Speaking} (TBS), an RG framework that trains the RG model to \emph{explicitly generate} the implicit knowledge and use this knowledge to generate a response, inspired by inquiry-based discovery learning~\cite{bruner1961act}.
We argue that this decomposition brings three major benefits: 
1) compared with end-to-end RG, generated knowledge \emph{augments} and/or \emph{constrains} RG to produce more informative responses; 
2) compared with knowledge-retrieval models, \emph{explicitly generating} intermediate groundings can potentially generalize to knowledge not included in KBs and synergize with the RG process;
3) explicitly generated implicit knowledge used in RG provides a faithful explanation of the response intent.

This new RG paradigm poses three main challenges: (1) how to \emph{identify} implicit commonsense knowledge associated with dialogue turns for training the knowledge generation module; 
(2) how to \emph{represent} structured knowledge in natural language (NL) for neural generative models; and (3) how to \emph{integrate} knowledge and dialogues while \emph{distinguishing} implicit and explicit parts in responses.
To collect knowledge associated with each dialogue instance for training the TBS generative model,
we propose weak supervision procedures to automatically align knowledge with each dialogue turn, rather than manually collecting human-annotations, which is expensive and unscalable.
This is achieved by using ConceptNet~\cite{speer2017conceptnet} as our knowledge base and different matching approaches to identify the implicit knowledge. 
We explore several ways to \emph{format} knowledge originally represented as structured triples into natural language so that RG models can adapt to the \emph{knowledge+response} generation task easily. We experiment with structured triples, triples converted to natural language, and a more colloquial question answering format. To ensure a smooth transition between knowledge and dialogues, we consider using special symbols or prompts as separators. 

To evaluate the TBS framework,
we introduce new evaluation protocols to cover different aspects of the system, including response quality, knowledge quality, and how TBS models leverage generated knowledge. We conduct extensive human evaluations for different variants of our training procedure. Our experimental results show that our models produce more informative, specific, and responses that make more common sense compared to end-to-end RG models and other knowledge-augmented models such as knowledge-selection. Knowledge quality analysis shows that at least 85\% of generated knowledge makes sense and is relevant, and the generated novel knowledge (not in ConceptNet) also has high quality. Furthermore, our TBS model even outperforms an RG model that takes in knowledge obtained using \emph{ground-truth} responses, showing that explicitly generating implicit knowledge is a promising direction for response generation in open domain dialogue systems. 




\vspace{0.3cm}
\section{Problem Formulation}\label{prob_definition}
Our TBS RG paradigm extends the traditional RG setting by incorporating an additional component of \emph{implicit knowledge} in the generation process to externalize the knowledge grounding step in RG.

\subsection{Response Generation}
We follow the common dialogue response generation setup~\cite{weizenbaum1966eliza, ritter-etal-2011-data, sordoni2015neural}: given a dialogue \emph{history} $H$ (a sequence of dialogue utterances), generate an appropriate \emph{response} $R$.
Current neural RG models often frame this task as a \emph{conditional language modeling} problem. Specifically, given a \emph{history ($H$)} consisting of a sequence of $n$ dialogue turns: $X_1, X_2, ..., X_n$ (each turn refers to an utterance containing a sequence of $t_i$ tokens: $x_{i,1}, x_{i,2}, ..., x_{i, t_i}$) and a \emph{response ($R$)} sentence $Y$ comprised of a sequence of $m$ tokens $y_1, y_2, ..., y_m$, RG models aim to learn the conditional probability distribution by training on human dialogues:
\begin{equation}
    P_{\theta}(R|H)=\prod_{i=1}^{m}P_{\theta}(y_i|y_{<i}, X_1,...,X_n).
\end{equation}

\subsection{Implicit Knowledge Generation}\label{2.2}
To make the implicit knowledge grounding step explicit, we introduce a new component to RG -- implicit knowledge that is \emph{conditioned on} the dialogue history $H$. We use $I$ to denote the implicit knowledge for brevity, which contains multiple natural language (NL) statements  $I=Z_1, Z_2, ...$ (each containing a sequence of tokens: $z_{i,1}, z_{i,2}, ...$) expressing commonsense knowledge. For example, in Figure~\ref{fig:motivation}, ``\emph{rose is a type of flower}'' and ``\emph{rose is a symbol of love}'' are two NL statements expressing the implicit commonsense knowledge. 
To emulate realistic conversation scenario, we also \emph{fuse} dialogue history $H$ in traditional RG with implicit knowledge $I$ for each turn and denote it with $H'$. i.e. $H'=X_1, I_1, X_2, I_2 ..., X_n$, where $I_i$ indicates the implicit knowledge statements for the i-th turn in the dialogue history. 

To externalize the knowledge grounding step, inspired by how humans communicate and inquiry-based learning~\cite{bruner1961act,shwartz-etal-2020-unsupervised}, our TBS RG paradigm requires models to first \emph{generate} implicit knowledge $I$ conditioned on $H'$, 
\emph{i.e.} $P_{\theta}(I_n|H'=X_1, I_1, X_2, I_2 ..., X_n)$.

\section{Learning to Generate Implicit Knowledge by Self-Talk}\label{method}
\begin{figure*}[t]
	\centering
	\includegraphics[width=0.9\linewidth]{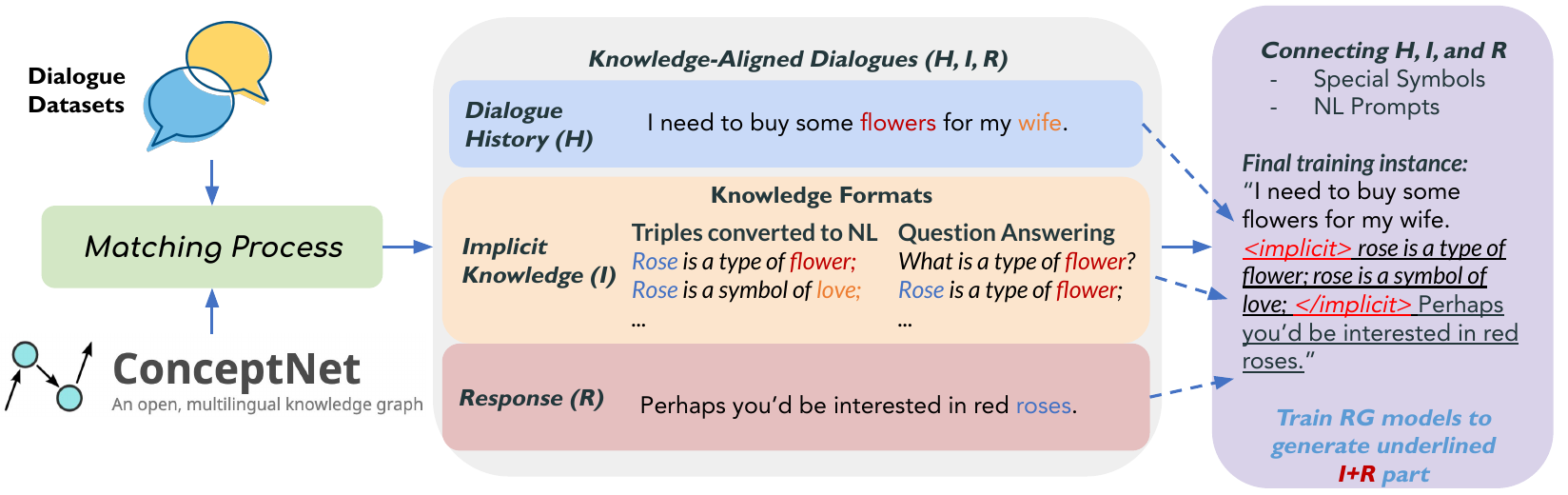}
	\caption{
	{\small \textbf{Method illustration.} We first propose matching approaches to construct \emph{knowledge-aligned dialogues}. Then we consider different alternatives to represent implicit knowledge. Finally, we connect knowledge and dialogue and ask models to generate both knowledge and responses given history.}}
	\label{fig:method}
\end{figure*}

This section introduces our proposed TBS method to train a generative model that can \emph{both} talk with itself to explicitly generate background commonsense knowledge ($P_{\theta}(I|H')$ ) and then generate response afterwards, $P_{\theta}(R|H', I)$. 
Figure~\ref{fig:method} illustrates the process to train the TBS models.
To pair each dialogue with appropriate implicit knowledge, we first define a \emph{matching process} and use  ConceptNet~\cite{speer2017conceptnet} as the implicit knowledge source (Section \ref{matching}).
Then, to construct training instances, we face two key method design choices: how to represent knowledge (\ref{representation}) and how to connect the knowledge with the dialogue (\ref{transition}).
Finally, we train TBS RG models to learn $P_{\theta}(I|H')$ and $P_{\theta}(R|H', I)$ with the same parameters $\theta$.
The following sections explain these components in details. 


\subsection{Knowledge-Aligned Dialogues}\label{matching}
To train TBS models we need dialogue datasets consisting of a dialogue history, a response, and the knowledge statement connecting them. 
We focus on two methods that create \emph{weakly-supervised knowledge labels} for dialogues as they are more scalable and cost less than human annotations. 

\paragraph{Hard-Matching} 

The hard-matching process first \emph{lemmatizes} all the non-stop words in each utterance, then it identifies 
knowledge triples 
whose two concepts appear in an utterance and the next turn respectively.  
This is the same as the filtering process in~\citet{zhou-etal-2021-commonsense} and is closely related to distant supervision methods for relation extraction~\cite{craven1999constructing,mintz2009distant}. For more details, refer to Appendix~\ref{appendix_matching}.



\paragraph{Soft-Matching Using Embedding Similarity}
Hard-matching only captures the surface form and neglects many important semantic relations between words. 
We thus develop a soft-matching procedure using embedding similarity from SentenceBERT~\cite{reimers-2019-sentence-bert} to measure semantic relations between dialogue turns and triples in ConceptNet. 
Specifically, we first extract candidate triples from ConceptNet with one concept appearing in the $i^{th}$ turn.
Next, we form a \emph{query} by concatenating the $i^{th}$ turn and the next $(i+1)^{th}$ turn response. 
Finally, we encode the query and all triple candidates using SentenceBERT and use cosine similarity to find the semantically closest triples as matched knowledge.  More details are presented in Appendix~\ref{appendix_matching}.

\subsection{Knowledge Representation}\label{representation}
Implicit commonsense knowledge $I$ stored in ConceptNet is in the form of \emph{(subject $s$, relation $r$, object $o$)} triples, such as \emph{(rose, TypeOf, flower)}, which is not compatible with RG models, which operate on NL sentences and may not include relation tokens in their trained vocabulary.
Here we design two alternatives to represent the grounded knowledge  and use the implicit knowledge in Figure~\ref{fig:motivation} as a running example.


\paragraph{Map Relations to Natural Language (NL)}
To convert ConceptNet triples into NL, we follow a common practice and map every relation $r$ in the triple to its NL template, and fill in $s$ and $o$ in the template~\cite{levy2017zero}. We use the same mapping as that used in COMET~\cite{bosselut-etal-2019-comet}, covering all standard types of relations in ConceptNet. For example, \emph{rose is a type of flower; rose is a symbol of love}. 

\paragraph{Information-Seeking Question-Answer Pairs}
Another format to convert triples to NL sentences is through asking and answering information-seeking questions. \citet{shwartz2020unsupervised} designed templates of information-seeking questions and answers to provide background knowledge for LMs. We adopt a similar strategy and design a template for each relation in ConceptNet. For example, \emph{What is a type of flower? Rose is a type of flower. Rose is a symbol of what? Rose is a symbol of love}. The mappings we use for these two types of representations are shown in Appendix~\ref{appendix_mapping}.

\subsection{Knowledge-Dialogue Transition}\label{transition}
To help our RG models learn the TBS paradigm and generate outputs structured similarly, i.e., implicit knowledge first and then responses, we need to properly connect knowledge and dialogues in our data. Here we consider two alternatives for creating such a transition.

\textbf{Special symbols}. Following the common practice of separating sequences in neural LMs~\cite{radford2018improving,devlin-etal-2019-bert}, we use a special symbol to serve as the separator. We enclose the implicit knowledge $I$ with special symbols ``$<$implicit$>$'' and ``$<$/implicit$>$'' and add it between $H'$ and $R$, for example, ``\emph{$<$speaker1$>$ I need to buy some flowers for my wife. $<$implicit$>$ rose is a type of flower $<$/implicit$>$ $<$speaker2$>$ Perhaps you'd be interested in red roses.''} 
    
\textbf{Natural language prompts}. More recent work has found that NL prompts help LMs to perform better on various downstream tasks, including natural language generation (NLG)~\cite{brown2020language, liu2021pre, zheng2021exploring}. Here we use the NL prompts to prompt RG models to \emph{generate} implicit knowledge and responses. We use ``\emph{The following background knowledge is helpful for generating the response:}'' to elicit knowledge and ``\emph{Grounded on the background knowledge, what does the speaker probably say in the next response?}'' to elicit response.

\subsection{Model Training}
After constructing knowledge-aligned dialogues, each of our data instances is a sequence of tokens with three components: a dialogue history $H'$ fused with potential implicit knowledge after each turn, implicit knowledge (empty or non-empty) $I$, and a response $R$. 
We split each instance $d(H', R, I) \in D$ to first train the model to generate just the knowledge $I$ based on $H'$, $P_{\theta}(I|H')$, and then train it to generate $R$ based on both $I$ and $H'$, $P_{\theta}(R|H', I)$. 

Formally, we follow standard way of modeling $P_{\theta}$ in auto-regressive neural RG models and use Maximum Likelihood Estimation (MLE) to train our model to maximize $P_{\theta}(I|H')$ (knowledge generation KG) by minimizing the conditional negative log-likelihood loss (NLL):
\begin{equation*}
    \mathcal{L}_{KG} = - \sum_{i=1}^{m}\log P_\theta(Z_i|Z_{<i},X_1,...,X_n),
\end{equation*}
 where $Z_i$ is the i-th statement in $I$. And to model $P_{\theta}(R|H', I)$ we minimize:
\begin{equation*}
    \mathcal{L}_{RG}  = - \sum_{i=1}^{m}\log P_\theta(y_i|y_{<i},X_1,I_1..., X_n).
\end{equation*}
We train one generative model on these losses in one-pass with splitted instances for KG and RG instead of multiple training phases. During inference, we only provide dialogue history as input and the model has to generate knowledge and responses.





\section{Experiment Setup} \label{setup}
\subsection{Dataset}


%
We consider dialogues from four datasets: DailyDialog~\cite{li2017dailydialog}, EmpatheticDialogues~\cite{rashkin2019towards}, MuTual~\cite{cui2020mutual}, and SocialIQA-prompted Commonsense-Dialogues ~\cite{zhou-etal-2021-commonsense}.
For training, we use the filtered version of the four datasets from~\citet{zhou-etal-2021-commonsense}, which ensures each dialogue contains at least one commonsense knowledge triple from ConceptNet. 
In total, the training data contains 31k dialogues with 159k utterances.
We reserve 10\% of data as a development set for evaluating model training and selecting hyper-parameters. 
Table~\ref{tab:data_stats} shows the number of instances resulted from applying our hard- and soft-matching procedures to our training data in order to construct knowledge-aligned dialogues. 

For testing dialogues, to not bias our evaluation toward where common sense is crucial in making the response, we use the test data from the \emph{original data distribution} of the 4 datasets mentioned above. The testing data consists of around 3k dialogues.


\begin{table}[]
\centering
\small
\resizebox{\columnwidth}{!}{
\begin{tabular}{c|c|c|c}
                  & \# Instances  & Avg \# turns     & Avg \# knowledge     \\ \hline
Dialogues-Only & 159k & 4.3 &  0 \\ 
Hard-match   &   57k        &     4.5      &     1.4      \\ 
Soft-match          &    71k   &     4.6     &  2.8  \\ \hline
\end{tabular}
}
\caption{\small Dialogue data statistics.}
\label{tab:data_stats}
\end{table}

\subsection{Compared Methods}\label{baselines}

We use DialoGPT-medium~\cite{zhang-etal-2020-dialogpt} as our base model, which is a commonly-used end-to-end RG model.
We fine-tune \textbf{DialoGPT} using all of the 159K dialogue instances. 
We also use DialoGPT to serve as the backbone model 
and consider three variables in our TBS model configuration introduced from Sections~\ref{matching} to~\ref{transition}: \textbf{hard}-matching or \textbf{soft}-matching, special \textbf{symbol} as separator or NL \textbf{prompt}, and triple-converted-\textbf{NL} to represent knowledge or information seeking \textbf{QA} pairs. To justify our choice of using one model to do both KG and RG, we also compare with \textbf{TBS-Two Model} where we train separate models for knowledge generation (KG) and RG using the same training data. Our default model configuration is \emph{hard-symbol-NL}.

We also compare several knowledge-grounded RG baselines that \emph{retrieve} external knowledge or \emph{generate} knowledge with another model.
For retrieval, we follow most common approaches in knowledge-selection~\cite{zhao2017learning, wolf-etal-2020-transformers, eric2021multi} and train RoBERTa~\cite{liu2019roberta} to classify triples using our knowledge-aligned data (matched or not matched), and use it to label candidate triples during testing (\textbf{KS-RoBERTa}). 
For the generative model, we use COMET~\cite{bosselut-etal-2019-comet} as a commonsense knowledge generator 
(\textbf{KG-COMET}). 

Furthermore, we consider RG models that take the hard-matched or soft-matched knowledge obtained from the \emph{ground-truth response} (\textbf{Hard-GT} and \textbf{Soft-GT}).
Note that though there is noise in hard-matching or soft-matching procedure, this setting uses the next turn response and is likely to provide relevant knowledge. Implementation details for all the models are shown in Appendix~\ref{appendix_implementation}. 



\subsection{Evaluation Protocol}\label{eval_protocol}


\paragraph{Automatic Evaluation}
We use standard natural language generation metrics such as BLEU~\cite{papineni2002bleu}, METEOR~\cite{banerjee2005meteor}, ROUGE~\cite{lin2004rouge}, CIDEr~\cite{vedantam2015cider} and SkipThoughts~\cite{kiros2015skip}. 
We also use GRADE~\cite{huang2020grade}, a reference-free metric shown to have consistent correlation with human judgements ~\cite{yeh2021comprehensive} to ensure the validity of experimental results.

\paragraph{Human Evaluation}
We conduct extensive human evaluation using 300 randomly sampled instances from unseen test dialogues described above.
For \textbf{response quality}, we conduct \emph{pairwise comparison} where we present a dialogue history and two responses made by two different models and ask them to choose one or select ``\emph{not sure}'' based on different criteria~\cite{zhou2018commonsense,zhang2020dialogpt}\footnote{We choose to conduct pairwise comparison since multiple previous work has shown that it produces a more reliable evaluation than directly asking humans to score the response, which is a highly subjective task~\cite{amidei2019use, callison-burch-etal-2007-meta, celikyilmaz2020evaluation}}.
We evaluate on \emph{six} dimensions: which response is more \emph{grammatical}, \emph{coherent}, \emph{engaging}, \emph{informative}, \emph{specific}, and \emph{makes common sense}~\cite{zhang2020dialogpt,roller2020recipes}.
More details of the instructions for annotators on each dimension with examples are included in  Appendix~\ref{appendix_evaluation}.
For \textbf{knowledge quality}, we evaluate the generated knowledge in isolation (``\emph{does this knowledge make sense}'') and in conjunction with the context for relevance.
We perform majority voting per instance using three annotators from Amazon Mechnical Turk (AMT). We use Fleiss' Kappa ($\kappa$) ~\cite{fleiss1971measuring} to measure agreement among the annotators.

\section{Results} \label{result}

\begin{table*}[tb]
\centering
\resizebox{\linewidth}{!}{
\begin{tabular}{l|c|c|c|c|c|c|c}
\multicolumn{1}{c|}{Model Variants} & Grammatical  & Coherent   & Engaging & Informative & Specific & Common Sense & \textbf{Avg} \\ \hline
TBS-\textbf{soft}-symbol-NL          & \textbf{53.0/10.0\%} & 46.3/8.7\%          & 48.7/9.3\%       & 41.7/20.6\%          & 51.7/6\%       & \textbf{52/7\%}  & 50.5/10.3\% \\
TBS-hard-\textbf{prompt}-NL              & 50.3/4\%           & 49/7.3\%         & 47/9\%          & 49.4/6\%          & 51/3\%         & 48.3/2.7\%      &   49.2/5.3\%   \\
TBS-hard-symbol-\textbf{QA}          & \textbf{53/6.7\%} & \textbf{53.6/5.6\%} & 51.3/4.7\%       & 51.3/3.7\%          & 51.3/5\%       & \textbf{54/3.7\%}  & \textbf{52.4/4.8\%}
\end{tabular}
}
\caption{\small Human evaluation on \textbf{response quality} when comparing different model variants. We show the percentage of times annotators prefer each variant to \emph{TBS-hard-symbol-NL} and ties, \emph{i.e.} wins/ties\%. Bold-faced numbers indicate statistical significance (p $<$ 0.05) improvement.
}
\label{tab:variants}
\end{table*}			
					
\begin{table*}[tb]
\centering
\resizebox{\linewidth}{!}{
\begin{tabular}{l|c|c|c|c|c|c|c|c|c}
\multicolumn{1}{c|}{Models}     & \textbf{GRADE} & BLEU-1         & BLEU-2         & BLEU-3         & BLEU-4         & METEOR         & ROUGE-L        & CIDEr          & SkipThoughts   \\ \hline
DialoGPT-ft~\cite{zhang2020dialogpt}                  & 0.704          & 0.060          & 0.026          & 0.013          & 0.007          & 0.061          & 0.076          & 0.087          & 0.700          \\ 
KS-SBERT~\cite{reimers-2019-sentence-bert}                 & 0.640          & 0.067         & 0.024          & 0.011          & 0.005          & 0.061          & 0.066         & 0.047          &	0.676         \\ 
KS-RoBERTa~\cite{eric2021multi}                 & 0.651          & 0.073          & 0.026          & 0.011          & 0.005          & 0.061          & 0.069          & 0.051          & 0.676          \\ 
KG-COMET~\cite{bosselut-etal-2019-comet}              &      0.648          &     0.080           &      0.032          &      0.015          &     0.007           &        0.069        &       0.076         &        	0.069        &        0.690        \\ 
\hline
\textit{TBS-Two Model} & 0.722          & \textbf{0.091}*          & 0.033          & 0.014          & 0.006 & 0.070          & 0.073          & 0.054         & 0.677 \\ 
\textit{TBS} & \textbf{0.739}** & \textbf{0.091}* & \textbf{0.037} & \textbf{0.020} & \textbf{0.012}          & \textbf{0.075}*          & \textbf{0.084}*          & \textbf{0.087}* & \textbf{0.703}        
\\ 
\hline
Hard-GT              & 0.702          & \textbf{0.091}          & 0.035          & 0.017          & 0.008          & \textbf{0.075} & \textbf{0.084} & 0.086          & 0.696          \\ 
Soft-GT              &       0.642        &       0.070         &      0.024          &       0.011         &       0.005         &       0.063        &       0.069         &        0.053        &    0.680            \\ 

\end{tabular}
} 
\caption{\small Automatic evaluations using multiple metrics on \textbf{response quality}. All models are based on DialoGPT-medium. Bold-faced are the best performance. One ``*'' indicates statistical significant (p $< 0.05$ in Wilcoxon signed-rank test) improvement upon the best-performing non-GT baseline and ``**'' indicates significant improvement upon the GT baselines.
}

\label{tab:automatic}
\end{table*}

\begin{figure*}[tb]
	\centering
	\includegraphics[width=1.0\linewidth]{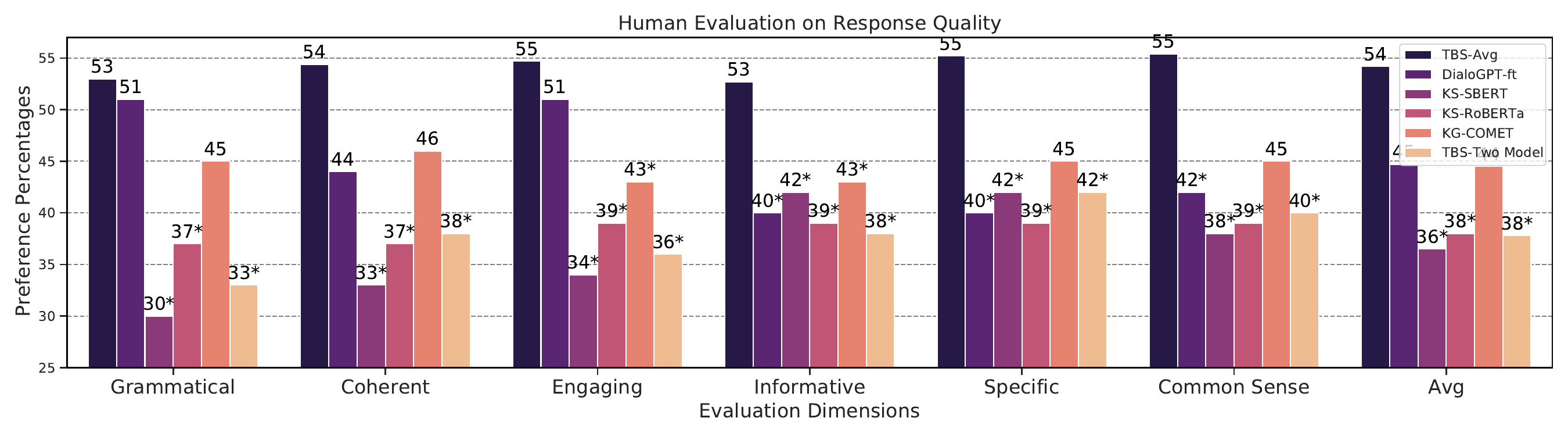}
	\caption{ \small
	\textbf{Human evaluation results} for pairwise comparison between TBS and a baseline. We show preference percentages for each model. ``*'' indicates statistical significance difference. For TBS we show averaged preferences.
	}
	\label{fig:human_eval}
\end{figure*}

\begin{figure}[tb]
	\includegraphics[width=1.15\columnwidth]{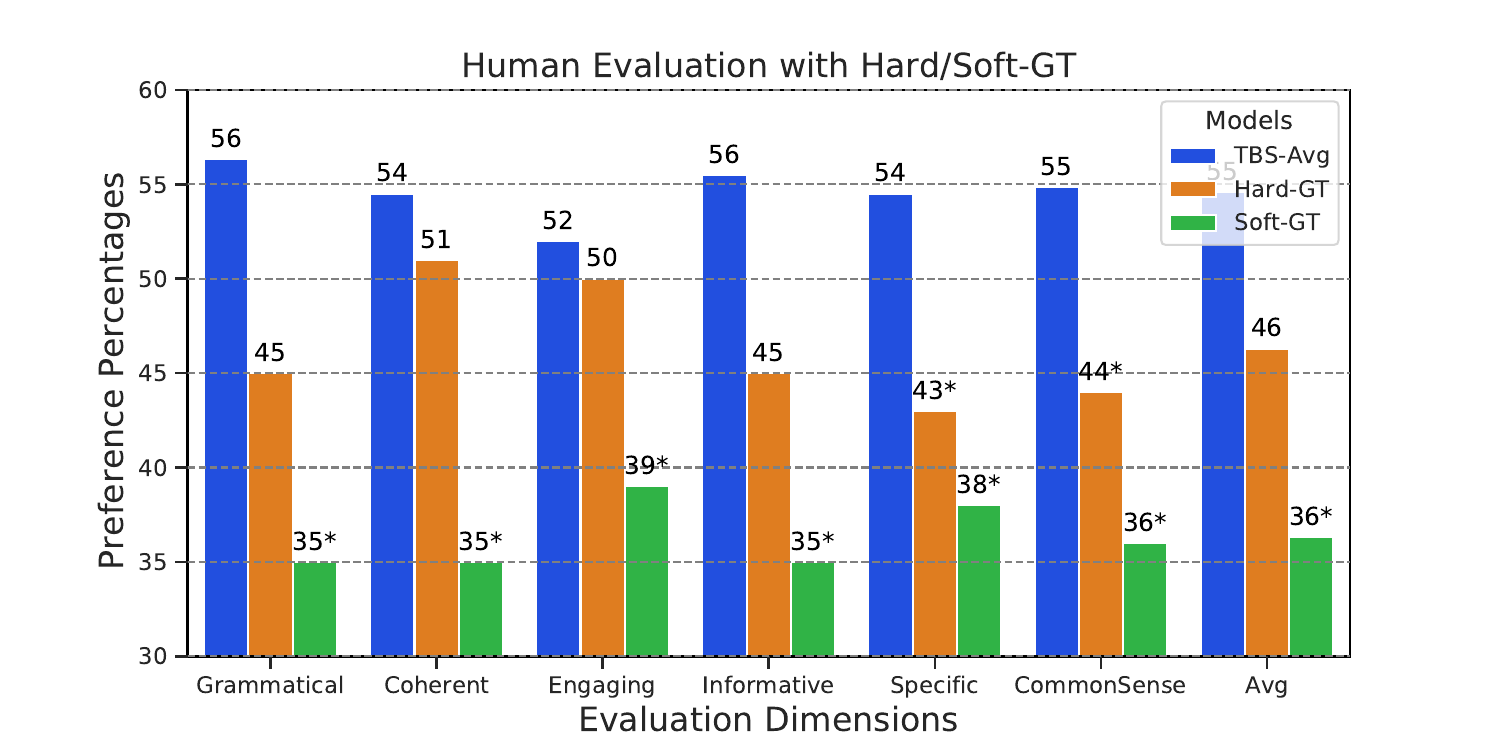}
	\caption{
	\small \textbf{Human evaluation} comparing TBS with models that have access to \emph{ground-truth} responses.	
	}
	\label{fig:human_eval_GT}
\end{figure}

By evaluating our TBS model variants with other baselines, we aim to address the following questions: 1) do TBS models produce better responses than standard end-to-end RG models? 2) compared with other approaches to retrieve or generate additional knowledge, is TBS more helpful for RG? 3) do TBS RG models generate knowledge that makes sense and is relevant to the dialogue context? 4) do TBS models faithfully leverage the generated knowledge?

\subsection{Performance of Response Generation}
\paragraph{Model variant analysis} 
To find the best-performing configuration of our TBS method, we consider alternatives as discussed in Sections~\ref{matching} to~\ref{transition}, and conduct 4 pairwise comparisons: \emph{soft} vs. \emph{hard}, \emph{prompt} vs. \emph{symbol}, and \emph{QA vs. relation-converted NL format}.
From Table~\ref{tab:variants}, we find that using soft-matching to create knowledge-aligned dialogue dataset produces more grammatical responses and responses that make more common sense, with $\kappa$=0.64-0.73, indicating substantial agreement according to one interpretation from~\citet{landis1977measurement}.
Using QA to represent knowledge makes the responses more grammatical, coherent, commonsensical, and also achieves the best performance on average on six dimensions. 
We also compare results that combine these alternatives, \emph{e.g., soft-symbol-QA} (due to space constraints, results are shown in Appendix~\ref{appendix_variants}), however, we do not observe significant improvements after combining these alternatives and our best configuration in terms of average improvement is still \emph{hard-symbol-QA}.
We thus use \emph{hard-symbol-QA} as our final configuration and refer to it as \emph{TBS} throughout this section. 

\paragraph{Does TBS produce better responses vs. end-to-end RG?} 
By comparing TBS and \textit{end-to-end} DialoGPT-ft model in Table~\ref{tab:automatic} and Figure~\ref{fig:human_eval}, we find that TBS models produce better-quality responses using both automatic and human evaluations.
Specifically, even though hard-matching only annotates about 33\% of the training instances,
TBS outperforms end-to-end RG model significantly on most automatic metrics.
From human evaluation ($\kappa$=0.62-0.69), we find our TBS model performs on par with DialoGPT trained on more data in grammar, coherence, and engagingness, and achieves statistically-significant (p$<$ 0.05) improvement on informativeness, specificity, and the common sense aspects of generated responses\footnote{We also conducted direct scoring in human evaluations and observed significant improvement (on average 7.3 out of 10 for TBS vs. 5.9 for DialoGPT-ft), but since it results in lower agreement ($\kappa$=0.49), we focus on comparative evaluation.}. We argue that by providing weakly-supervised knowledge labels and TBS training, RG models require less data and can generate quality responses with improvement in the informativeness, specificity, and common sense aspects of the responses.

\paragraph{Is TBS knowledge generation better than other knowledge-augmented RG?}
We compare TBS models with other knowledge-augmented baselines that retrieve knowledge from ConceptNet using embedding scores (KS-SBERT) or a trained selector (KS-RoBERTa), or generate from \emph{another} model (KG-COMET). From Table~\ref{tab:automatic}, we find that these models perform similarly to the end-to-end DialoGPT model and are outperformed by TBS models on most automatic metrics. Figure~\ref{fig:human_eval} shows that while TBS methods have significant improvements on all dimensions against knowledge-selection baselines, COMET as a knowledge generator has smaller gaps on informativeness, specificity, and common sense, but is outperformed significantly on grammar, coherence, and engagingness.

Next we compare against the setup where we feed the model the knowledge that is derived using the \emph{ground-truth} response (Hard/Soft-GT), \emph{i.e.}, the provided knowledge is obtained using concepts appearing in the ground-truth response.
From Table~\ref{tab:automatic}, we surprisingly find that even though our proposed TBS model has no access to response-leaking knowledge labels and is trained on much less data, the TBS RG model still achieves statistically significant improvement on GRADE and BLEU-4. And from human evaluation results in Figure~\ref{fig:human_eval_GT}, TBS model significantly improves the specificity and common sense aspect of responses while stays on par on other evaluation dimensions compared with the hard-GT model and improves even more compared with soft-GT. 
We find that one potential explanation is that only around 55\% of Hard-GT knowledge is labeled as \emph{used in response} whereas it is 77\% in our TBS model (see Section~\ref{analysis}). 
This is also related to how the RG model leverages the knowledge in training.  
Further analysis is needed to understand the effect of knowledge and the relationship between knowledge and responses.  
\begin{table}[tb]
\centering
\small
\resizebox{\columnwidth}{!}{
\begin{tabular}{l|c|c|c} 
\textbf{Model}          & \textbf{Novel}          & \textbf{Makes Sense}  & \textbf{Relevant}        \\ \hline
KS-SBERT & 0\%   &    91.7\%*       &     85.0\%              \\
KS-RoBERTa  & 0\%   &   77.7\%*              &  76.3\%           \\
KG-COMET &  63.3\%  &  68.3\%/63.2\%          &  67.5\%/68.9\%                  \\ \hline
TBS-two-model     &   46.3\%   &    89.0\%/85.6\%      &  90.7\%/90.2\%              \\
TBS-one-model      & 44\%      &  86.3\%/85.9\%            & 85.7\%/86.5\%             \\  \hline       
\end{tabular}
} 
\caption{ \small Human evaluation on \textbf{knowledge quality}. For models that generate novel (not in ConceptNet) knowledge, we show \emph{non-novel/novel} percentages. ``*'' means knowledge is from ConceptNet (not generated).
}
\label{tab:KQ}
\end{table}

\subsection{Quality of Generated Knowledge }
We then examine how well TBS RG models learn to generate knowledge on unseen dialogues. We use human evaluation and focus on three dimensions: does the model generate \emph{novel} knowledge that does not appear in ConceptNet? does the generated knowledge statement \emph{make sense} as a standalone fact? and is the generated knowledge \emph{relevant} to the dialogue context? For the first question we directly query from ConceptNet and show percentages. 
For the latter two we follow Section~\ref{eval_protocol} and show the percentages that MTurkers think the knowledge makes sense and is relevant from the 300 sampled test instances (the same used in response quality).
We test our TBS model, the two-model variant, and other knowledge-augmented baselines introduced in Section~\ref{baselines}. 

\paragraph{Around 85\% of knowledge generated from TBS makes sense and is relevant}
Table~\ref{tab:KQ} shows that TBS models can generate implicit knowledge that makes sense and is relevant to the context for around 85\% of the time as judged by human annotators ($\kappa$=0.73-0.80). Compared with knowledge-selection models that retrieve knowledge from ConceptNet, TBS generates knowledge that is similar in terms of common sense and has better relevance to the dialogue history. Compared with COMET that also generates knowledge, we find TBS models generate more knowledge that follows common sense and is relevant to the dialogue. Comparing two-model and one-model TBS, we find that two-model generates more knowledge that makes sense and is relevant, although its response quality is poorer (Table~\ref{tab:automatic} and Figure~\ref{fig:human_eval}). This might be due to model synergies when learning both knowledge generation and response generation.

\paragraph{Model generates \emph{novel} knowledge}
We find a significant portion of \emph{novel} knowledge generated from the COMET and TBS models that is not present in the training data. Furthermore, the quality of the generated novel knowledge is similar to that of knowledge existing in ConceptNet. COMET generates more new knowledge but the quality (both common sense and relevance) is significantly lower than TBS models. We include some examples of novel knowledge generated in Appendix~\ref{appendix_results}. In general we find that the new knowledge is complimentary to ConceptNet, not just a paraphrased version of existing triples (since in those cases the model will directly generate the ConceptNet triple). This shows a promising sign that TBS RG models can potentially generate good-quality novel knowledge labels for unseen dialogues.

\begin{table*}[tb]
\centering
\resizebox{\linewidth}{!}{
\small
\begin{tabular}{|cc|l|l|}
\hline
\multicolumn{2}{|c|}{\textbf{Dialogue History}}                                     & \begin{tabular}[c]{@{}l@{}}A: My last day of work was yesterday. \\ I got so emotional driving home and I cried. \\ B: I'm sorry to hear you were upset! \\ Why did you have to quit? \\ A: I'm going back to school full time. \\ I was there for more than 10 years. I will miss everyone.\end{tabular} & \begin{tabular}[c]{@{}l@{}}A: I am calling friends to let you know that\\  I am moving out of state. \\ B: What happened? Why? \\ A: I got a new job.\end{tabular} \\ \hline
\multicolumn{2}{|c|}{\multirow{2}{*}{\textbf{Human Response}}}                      & \multirow{2}{*}{\begin{tabular}[c]{@{}l@{}}B: Wow that is a long time! Hopefully you can stay \\ in touch with all of your former coworkers.\end{tabular}}                                                                                                                                                & \multirow{2}{*}{B: It must pay well for you to move.}                                                                                                              \\
\multicolumn{2}{|c|}{}                                                              &                                                                                                                                                                                                                                                                                                           &                                                                                                                                                                    \\ \hline
\multicolumn{2}{|c|}{\textbf{DialoGPT-ft}}                                          & B: That's a long time. I bet you'll miss your friends.                                                                                                                                                                                                                                                    & B: I am not sure where I want to live.                                                                                                                             \\ \hline
\multicolumn{1}{|c|}{\multirow{2}{*}{\textbf{TBS}}}        & Gen. Knowledge         & \textit{What is school related to? school is related to college;}                                                                                                                                                                                                                                         & \textit{\begin{tabular}[c]{@{}l@{}}What is job related to? job is related to work;\\ What is pay related to? pay is related to job;\end{tabular}}                  \\ \cline{2-4} 
\multicolumn{1}{|c|}{}                                     & Response               & \begin{tabular}[c]{@{}l@{}}B: I'm sorry to hear that. \\ I hope you have a good college experience!\end{tabular}                                                                                                                                                                                          & B: Oh wow! Will they pay you a lot?                                                                                                                     \\ \hline
\multicolumn{1}{|c|}{\multirow{2}{*}{\textbf{KS-RoBERTa}}} & Retr. Knowledge        & \textit{\begin{tabular}[c]{@{}l@{}}year is related to book; angry is a synonym of upset; \\ cry is related to upset;\end{tabular}}                                                                                                                                                                        & \textit{\begin{tabular}[c]{@{}l@{}}move is an antonym of easy; move is related to leave; \\ move is related to change;\end{tabular}}                               \\ \cline{2-4} 
\multicolumn{1}{|c|}{}                                     & Response               & B: You're not alone in that. I'm sorry you had to go through that.                                                                                                                                                                                                                                        & B: I am sure you will be able to make it work.                                                                                                                     \\ \hline
\multicolumn{1}{|c|}{\multirow{2}{*}{\textbf{KG-COMET}}}   & Gen. Knowledge         & \textit{I will miss everyone causes sadness}                                                                                                                                                                                                                                                              & \textit{I got a new job causes work}                                                                                                                               \\ \cline{2-4} 
\multicolumn{1}{|c|}{}                                     & Response               & B: I'm sure you will. I'm sure you'll do great.                                                                                                                                                                                                                                                           & B: I'm so sorry!  I hope you are able to get a new job!                                                                                                            \\ \hline
\multicolumn{1}{|c|}{\multirow{2}{*}{\textbf{Hard-GT}}}    & Hard-Matched Knowledge & \textit{\begin{tabular}[c]{@{}l@{}}year is related to long; long is related to miss; \\ long is related to full\end{tabular}}                                                                                                                                                                             & \textit{pay is related to job}                                                                                                                                     \\ \cline{2-4} 
\multicolumn{1}{|c|}{}                                     & Response               & B: I know the feeling. I'm going back to school too.                                                                                                                                                                                                                                                      & B: I am sorry to hear that. I hope you find a better job.                                                                                                          \\ \hline
\end{tabular}
}
\caption{Randomly sampled qualitative examples from our TBS model (TBS-hard-symbol-QA).
}
\label{tab:qual_examples}
\end{table*}

\begin{table*}[tb]
\centering
\resizebox{\linewidth}{!}{
\begin{tabular}{|l|l|l|}
\hline
\multicolumn{1}{|c|}{\textbf{Dialogue History}}                                                                                                                                                                                                                                                                                                                                                 & \multicolumn{1}{c|}{\textbf{Generated Knowledge}}                                                                                                                              & \multicolumn{1}{c|}{\textbf{Generated Response}}                                                                                                                   \\ \hline
\begin{tabular}[c]{@{}l@{}}\textless{}speaker1\textgreater My last day of work was yesterday. \\ I got so emotional driving home and I cried. \\ \textless{}speaker2\textgreater I'm sorry to hear you were upset! \\ Why did you have to quit? \\ \textless{}speaker1\textgreater Im going back to school full time. \\ I was there for more than 10 years. I will miss everyone.\end{tabular} & What is school related to? school is related to college;                                                                                                                       & \begin{tabular}[c]{@{}l@{}}\textless{}speaker2\textgreater I'm sorry to hear that. \\ I hope you have a good college experience!\end{tabular}                      \\ \hline
\begin{tabular}[c]{@{}l@{}}\textless{}speaker1\textgreater I am calling friends to let you know that\\  I am moving out of state. \\ \textless{}speaker2\textgreater What happened? Why? \\ \textless{}speaker1\textgreater I got a new job.\end{tabular}                                                                                                                                       & \begin{tabular}[c]{@{}l@{}}What is job related to? job is related to work; \\ What is pay related to? pay is related to job;\end{tabular}                                      & \textless{}speaker2\textgreater Oh wow! How long will you be working there?                                                                                        \\ \hline
\begin{tabular}[c]{@{}l@{}}\textless{}speaker1\textgreater There was something moving in the woods. \\ I looked out of the hunting shack and saw movement so I grabbed my gun.\end{tabular}                                                                                                                                                                                                     & What is move related to? move is related to change;                                                                                                                            & \textless{}speaker2\textgreater Did you change your mind?                                                                                                          \\ \hline
\begin{tabular}[c]{@{}l@{}}\textless{}speaker1\textgreater I love all of you guys! This was the most\\  special performance that I have ever had because of you all!\end{tabular}                                                                                                                                                                                                               & What is happy related to? happy is related to special;                                                                                                                         & \textless{}speaker2\textgreater Awww, you're so happy. It's always nice to see a happy couple.                                                                     \\ \hline
\begin{tabular}[c]{@{}l@{}}\textless{}speaker1\textgreater Two years ago, I discovered that my now ex-wife \\ was cheating on me with someone she worked with. \\ When I first discovered it, i was angry and hurt. \\ \textless{}speaker2\textgreater I bet you were, I did that to my husband\\  when I was young. I thought it was the solution to relief from abuse.\end{tabular}           & \begin{tabular}[c]{@{}l@{}}What is man related to? man is related to young; \\ What is young? young is a man; \\ What is man related to? man is related to woman;\end{tabular} & \begin{tabular}[c]{@{}l@{}}\textless{}speaker1\textgreater Yeah, I was so angry and hurt, \\ I thought I was going to be physically ill or something.\end{tabular} \\ \hline
\end{tabular}
}
\caption{Randomly sampled qualitative examples from TBS-hard-symbol-QA. We can find that  in most cases the generated knowledge makes sense and responses are grounded in the knowledge and not generic. However, we do find some error cases such as the last example the response does not use the knowledge. 
}
\label{tab:more_examples}
\end{table*}

\subsection{Performance Analysis}\label{analysis}

\paragraph{Most responses are knowledge grounded} 
To examine how TBS methods leverage knowledge for RG, we also present annotators a history, generated knowledge, and generated response, and ask them whether the knowledge is \emph{used in response}.
We find that around 77\% of generated knowledge is used in the generated response, \emph{i.e.}, the response is \emph{grounded} in the knowledge generated from TBS.

\begin{figure}[tb]
	\centering

	\includegraphics[width=0.9\columnwidth]{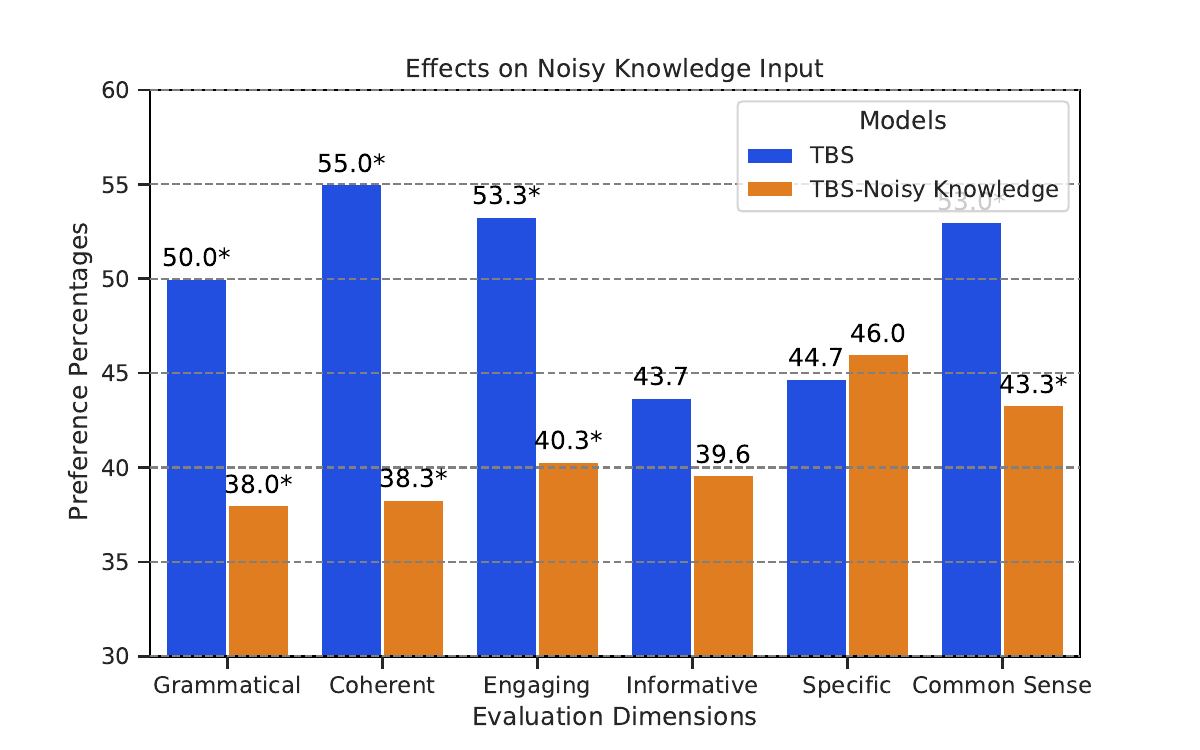}

	\caption{
	\small
	Effects of \textbf{noisy knowledge} on response quality. 
	}

	\label{fig:noisy_effects}
\end{figure}

\paragraph{Noisy knowledge heavily impacts quality}
To better showcase the connection between knowledge and response, we examine how knowledge quality generated from TBS methods can affect response quality. During inference, we randomly sample \emph{noisy knowledge} from another dialogue, feed it to the model to generate a response conditioned on irrelevant knowledge, and compare the response quality with response generated from TBS knowledge. 
Fig~\ref{fig:noisy_effects} shows that there is a statistically significant (p $\leq$ 0.05) drop in response quality in four dimensions. This indicates that the quality of knowledge input heavily influences response quality and that TBS models generate better responses because of its decent knowledge quality.

\paragraph{Qualitative examples and limitations}
We show several qualitative examples from different models and human responses in Table~\ref{tab:qual_examples}. We find that TBS generates relevant knowledge and responses grounded properly in that knowledge, whereas KS/KG models retrieve noisy knowledge and Hard-GT generates response not grounded in knowledge.

Here we present a summary of error patterns of TBS models and discuss potential directions to improve. More examples can be found in Table~\ref{tab:more_examples}. 
First, our matching procedures do not concern multi-hop triples that might be needed for complex reasoning chains. Second, ConceptNet mostly contains taxonomic and lexical knowledge (``\emph{RelatedTo, IsA, etc}''), limiting the diversity of generated knowledge from TBS models. We plan to explore other knowledge resources such as ATOMIC2020~\cite{Hwang2021COMETATOMIC2O} in the future. Third, currently the model always generates implicit knowledge. In future work, we are interested in training RG models that understand \emph{when} implicit knowledge is needed based on the dialogue context.

\section{Related Work}\label{rel_work}



\paragraph{Open-Domain Dialogue Generation}
Recent work focused on fine-tuning large pre-trained transformer models~\cite{radford2019language,zhang-etal-2020-dialogpt,roller2020recipes} on massive dialogue data. Knowledge-augmented RG has been studied extensively to alleviate the issue of generic or hallucinated responses~\cite{serban2017hierarchical,welleck2019neural, roller2020recipes}. Most work retrieves relevant knowledge from knowledge candidates (wikipedia or KBs) and generates responses after incorporating additional knowledge in dialogue context~\cite{ghazvininejad2018knowledge,zhou2018commonsense,wu2020diverse}. More recent work also explored other ways of constructing knowledge, such as by considering knowledge as a latent variable~\cite{tuan2020knowledge,li2020zero} and generating it implicitly. Our TBS framework differs from these two lines of work in that it \emph{explicitly generates} knowledge in text and uses one generative model for both knowledge generation and RG. 

\paragraph{Generating Knowledge for Natural Language Understanding (NLU)}
Although explicit knowledge generation (KG) for RG has not been explored, similar methods have been proposed for NLU tasks such as question answering~\cite{shwartz2020unsupervised}. Previous work has also explicitly generated \emph{rationales} that can be seen as helpful additional knowledge~\cite{rajani2019explain}. TBS differs from such work in that we consider a generative task and use the same generative model to do both KG and RG.



\section{Conclusion}\label{conclusion}
Inspired by how humans contribute to the common ground during communication, 
We propose to train RG models that explicitly generate implicit knowledge and then respond (TBS). 
This brings us three main benefits compared with prior end-to-end RG models: 1) more informative and coherent responses by augmenting with knowledge; 2) generated knowledge provides faithful explanations of RG model's inner-workings; 3) models do not rely on external knowledge bases in response generation time.
We first identify implicit knowledge in dialogues, explore different knowledge representation and transition choices, and demonstrate promising results compared with end-to-end and knowledge-grounded RG models from extensive evaluations.
We find strong and promising results for TBS RG model compared with end-to-end RG. In particular, TBS can produce good quality and novel knowledge, outperform end-to-end RG models despite training on less data, and even produce better responses than RG models that take ground-truth knowledge.
We hope our findings encourage more future studies on making RG models better emulate human communication process and produce better-quality responses.

\section*{Ethics and Broader Impact}
Our work aims to train RG models that explicitly generate implicit knowledge before responding. ~\citet{sheng2021nice} have found biases in DialoGPT (our base model) responses and ~\citet{mehrabi2021lawyers} have found representational harms in common sense resources. We acknowledge that the generated responses from our models might contain biases.
All of the dialogue datasets and models are in English, which benefits English speakers more. We have conducted human evaluation using Amazon Mechanical Turks. We pay turkers around \$15 per hour, well above the highest state minimum wage and engage in constructive discussions if they have concerns about the process. We also give each annotation instance enough time so that we do not pressure annotators.

\section*{Acknowledgments}
We thank anonymous reviewers for providing insightful feedback and members from Amazon Alexa AI team and INK and JAUNTS lab from USC. Pei Zhou, Jay Pujara, and Xiang Ren’s work on this project was funded by the Defense Advanced Research Projects Agency with award N660011924033. The research was also supported by gifts from Google.

\bibliographystyle{acl_natbib}
\bibliography{custom}

\clearpage
\appendix
\label{appendix}

\section{TBS Framework Details}
\subsection{Matching Detail}\label{appendix_matching}
\paragraph{Hard-Matching}
This process follows that used in ~\citet{zhou-etal-2021-commonsense}. We first identify potential candidates for concepts in ConceptNet~\cite{speer2017conceptnet}. For each utterance, we use a part-of-speech (POS) tagger to find the nouns, verbs, and adjectives that are not stopwords and then construct a set of potential concepts by including the lemmatized version of these words. The POS tagger, lemmatizer, and stopword list are from the Natural Language Toolkit (NLTK) package~\cite{bird2009natural}.
This step results in a set of concept words for \emph{each turn} of a dialogue. 

With a set of concepts we extract for every dialogue turn, we then identify a list of candidate triples $(e_1, r, e_2)$. We use the ConceptNet containing single-word concepts pre-processed by~\citet{zhou2018commonsense}. For each concept we identified in a turn, we store all triples in ConceptNet that contain this concept, either as subject or object. 

After getting a list of commonsense triples $(e_1, r, e_2)$ containing concepts in a particular turn using ConceptNet, we next examine if any of the \emph{other} entity in the triples appears in the concept set of the next turn. 
If we find such a match, 
we record this triple to be a commonsense assertion that might be implied in the response. 

\paragraph{Soft-Matching}
We reuse the first several steps of hard-matching to find a set of candidate triples for each dialogue turn, then instead of searching for the exact words in the next turn, we use embedding similarity from SentenceBERT~\cite{reimers-2019-sentence-bert} (specifically the ``\emph{all-MiniLM-L6-v2}'' variant, which is claimed to be a ``All-round model tuned for many use-cases. Trained on a large and diverse dataset of over 1 billion training pairs'')\footnote{\url{https://www.sbert.net/docs/usage/semantic_textual_similarity.html}}.

To select the final matched knowledge, we choose the top 3 triples from ConceptNet with the highest similarity. After examining the distribution of embedding similarities from SBERT, we also require the similarity to be above 0.4 to be matched to ensure quality matching.

\subsection{Mappings}\label{appendix_mapping}
We show complete mappings of relations from ConceptNet for both relation-converted NL and information-seeking QA pairs in Table~\ref{tab:mappings}.

\begin{figure}[]
	\centering
	\includegraphics[width=\columnwidth]{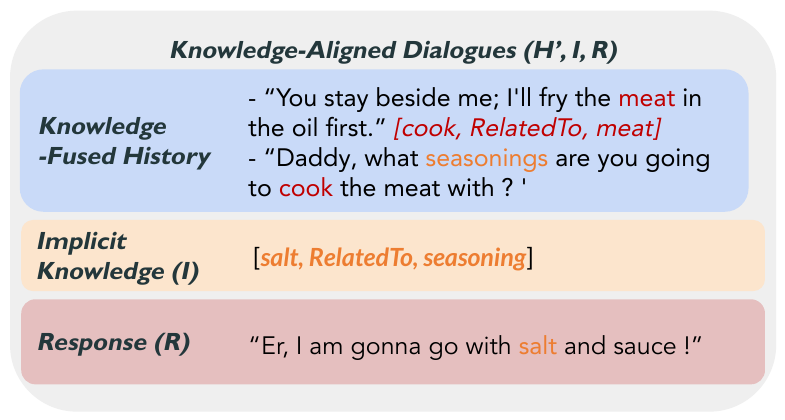}
	\caption{
	{\textbf{Data example.} We align implicit knowledge from ConceptNet~\cite{speer2017conceptnet} between dialogue turns and form each instance in three components.}
	}
	\label{fig:data}
\end{figure}

\section{Experimental Details}
\subsection{Implementation Details}\label{appendix_implementation}

We use base models from HuggingFace\footnote{DialoGPT-medium:~\url{https://huggingface.co/microsoft/DialoGPT-medium}} and implement TBS based on TransferTransfo~\cite{wolf2019transfertransfo}\footnote{\url{https://github.com/huggingface/transfer-learning-conv-ai}}. 
We fine-tune the model for 3 epochs with batch size 4 and set the learning rate to be 6.25e-5. We perform gradient accumulation for 8 steps and gradient clipping with a max norm of 1.0 and optimize using the Adam optimizer. For decoding, we use top-p nucleus sampling~\cite{holtzman2019curious} with temperature T (p = 0.9 and T = 0.7), and a maximum decoding length of 300 tokens. Note that since we are also generating knowledge, this maximum length is larger than normal RG models.
Our TBS models are mostly trained on 4 Quadro RTX 8000 GPUs and take around 5 hours. For automatic metrics, we use the nlg-eval package\footnote{\url{https://github.com/Maluuba/nlg-eval}} and the GRADE repo\footnote{\url{https://github.com/li3cmz/GRADE}}.



\subsection{Evaluation Detail}\label{appendix_evaluation}
\begin{figure*}[tb]
	\centering
	\includegraphics[width=1.0\linewidth]{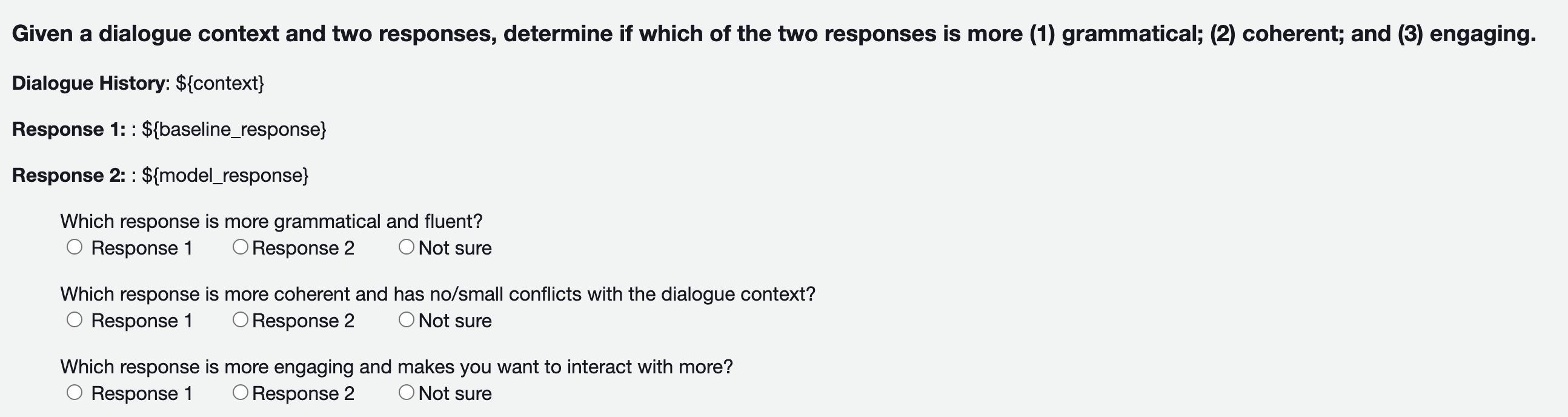}
	\caption{
	\textbf{Human evaluation interface} for response quality on dimensions: grammar, coherence, and engagingness.
	}
	\vspace{-0.1cm}
	\label{fig:turking_gce}
\end{figure*}

\begin{figure*}[tb]
	\centering
	\includegraphics[width=1.0\linewidth]{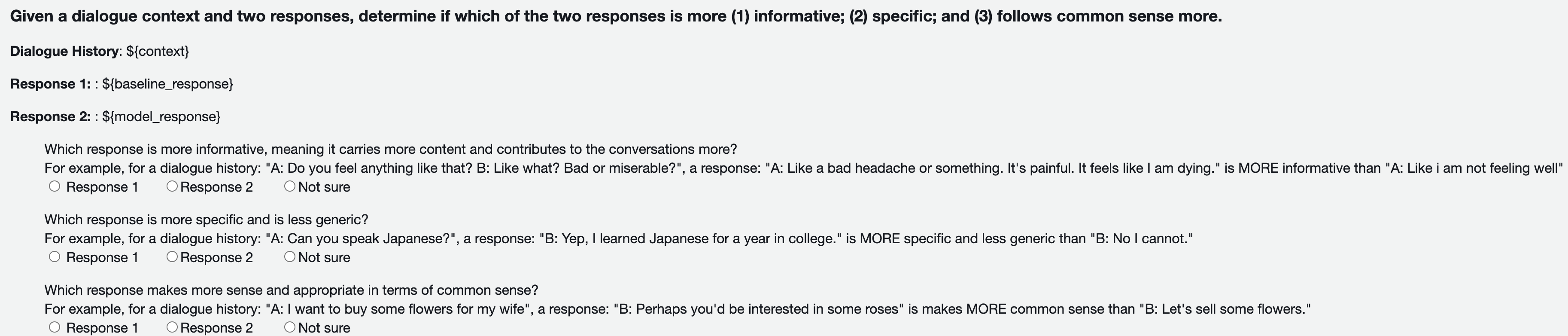}
	\caption{
	\textbf{Human evaluation interface} for response quality on dimensions: informativeness, specificity, and common sense.
	}
	\vspace{-0.1cm}
	\label{fig:turking_ics}
\end{figure*}

\begin{figure*}[tb]
	\centering
	\includegraphics[width=1.0\linewidth]{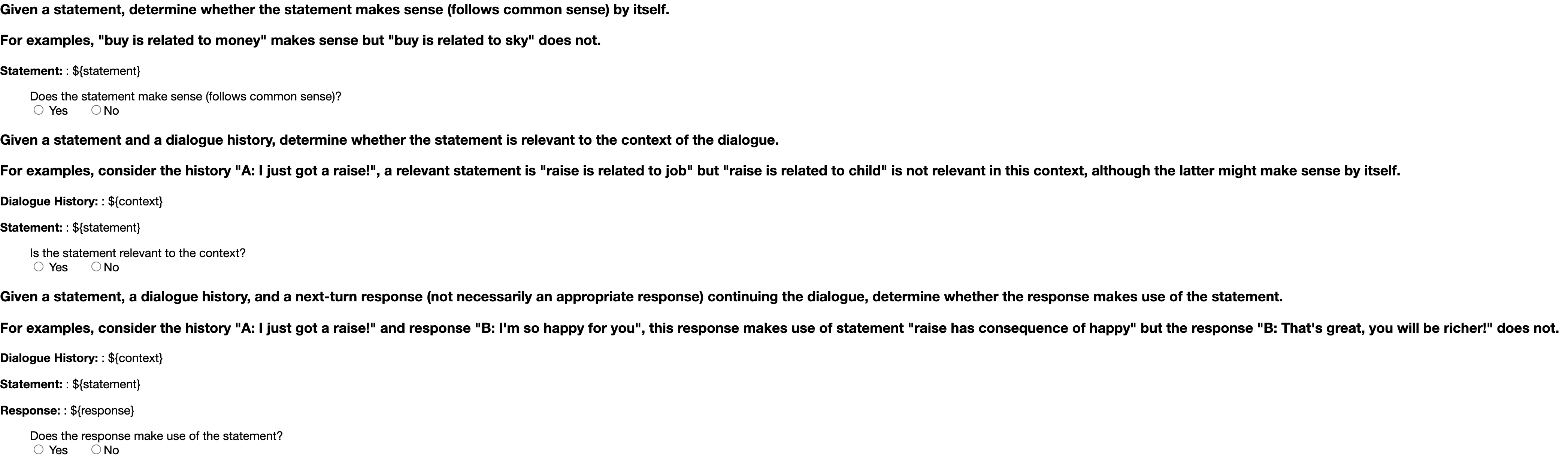}
	\caption{
	\textbf{Human evaluation interface} for knowledge quality with 3 questions: does the knowledge make sense as a standalone fact, is the knowledge relevant to the context, and does the generated resposne use the knowledge?
	}
	\vspace{-0.1cm}
	\label{fig:turking_kq}
\end{figure*}

We present the MTurk interface we use for response quality and knowledge quality evaluation in Figures~\ref{fig:turking_gce},~\ref{fig:turking_ics}, and~\ref{fig:turking_kq} including instructions and examples. We require turkers to have at least 500 numbers of HITs approved, with approval rate higher than 95\%, and from either Canada, UK, or US since our data is in English.

\section{Additional Results}\label{appendix_results}

\begin{table*}[tb]
\centering
\resizebox{\linewidth}{!}{
\begin{tabular}{|l|l|l|}
\hline
\textbf{Relation in ConceptNet} & \textbf{Relation-Converted NL} & \textbf{Information-Seeking QA}                                                                                                                                     \\ \hline
DefinedAs                       & is defined as                  & What is \textless{}concept1\textgreater defined as? | \textless{}concept1\textgreater is defined as \textless{}concept2\textgreater{}                               \\ \hline
DesireOf                        & desires                        & What does \textless{}concept1\textgreater desire of? | \textless{}concept1\textgreater desires \textless{}concept2\textgreater{}                                    \\ \hline
HasA                            & has a                          & What does \textless{}concept1\textgreater have? | \textless{}concept1\textgreater has \textless{}concept2\textgreater{}                                             \\ \hline
HasFirstSubevent                & starts with                    & What does \textless{}concept1\textgreater start with? | \textless{}concept1\textgreater starts with \textless{}concept2\textgreater{}                               \\ \hline
HasLastSubevent                 & ends with                      & What does \textless{}concept1\textgreater end with? | \textless{}concept1\textgreater ends with \textless{}concept2\textgreater{}                                   \\ \hline
HasPrerequisite                 & requires                       & What does \textless{}concept1\textgreater require? | \textless{}concept1\textgreater requires \textless{}concept2\textgreater{}                                     \\ \hline
HasProperty                     & has the property               & What property does \textless{}concept1\textgreater have? | \textless{}concept1\textgreater is \textless{}concept2\textgreater{}                                     \\ \hline
HasSubevent                     & requires                       & What subevent does \textless{}concept1\textgreater have? | \textless{}concept1\textgreater has subevent of \textless{}concept2\textgreater{}                        \\ \hline
IsA                             & is a                           & What is \textless{}concept1\textgreater{}? | \textless{}concept1\textgreater is a \textless{}concept2\textgreater{}                                                 \\ \hline
MadeOf                          & is made of                     & What is \textless{}concept1\textgreater made of? | \textless{}concept1\textgreater is made of \textless{}concept2\textgreater{}                                     \\ \hline
MotivatedByGoal                 & is motivated by                & What is \textless{}concept1\textgreater motivated by? | \textless{}concept1\textgreater is motivated by \textless{}concept2\textgreater{}                           \\ \hline
NotCapableOf                    & is not capable of              & What is \textless{}concept1\textgreater not capable of? | \textless{}concept1\textgreater is not capable of \textless{}concept2\textgreater{}                       \\ \hline
NotDesires                      & does not desire                & What does \textless{}concept1\textgreater not desire? | \textless{}concept1\textgreater does not desire \textless{}concept2\textgreater{}                           \\ \hline
NotHasA                         & does not have a                & What does \textless{}concept1\textgreater not have? | \textless{}concept1\textgreater does not have a \textless{}concept2\textgreater{}                             \\ \hline
NotHasProperty                  & does not have the property     & What property does \textless{}concept1\textgreater not have? | \textless{}concept1\textgreater does not have \textless{}concept2\textgreater{}                      \\ \hline
NotIsA                          & is not a                       & What \textless{}concept1\textgreater is not? | \textless{}concept1\textgreater is not a \textless{}concept2\textgreater{}                                           \\ \hline
NotMadeOf                       & is not made of                 & What is \textless{}concept1\textgreater not made of? | \textless{}concept1\textgreater is not made of \textless{}concept2\textgreater{}                             \\ \hline
PartOf                          & is part of                     & What is \textless{}concept1\textgreater a part of? | \textless{}concept1\textgreater is a part of \textless{}concept2\textgreater{}                                 \\ \hline
RelatedTo                       & is related to                  & What is \textless{}concept1\textgreater related to? | \textless{}concept1\textgreater is related to \textless{}concept2\textgreater{}                               \\ \hline
SymbolOf                        & is a symbol of                 & What is \textless{}concept1\textgreater a symbol of? | \textless{}concept1\textgreater is a symbol of \textless{}concept2\textgreater{}                             \\ \hline
UsedFor                         & is used for                    & What is \textless{}concept1\textgreater used for? | \textless{}concept1\textgreater is used for \textless{}concept2\textgreater{}                                   \\ \hline
AtLocation                      & is located at                  & Where is \textless{}concept1\textgreater{}? | \textless{}concept1\textgreater is located at \textless{}concept2\textgreater{}                                       \\ \hline
CapableOf                       & is capable of                  & What is \textless{}concept1\textgreater capable of? | \textless{}concept1\textgreater is capable of \textless{}concept2\textgreater{}                               \\ \hline
Causes                          & causes                         & What does \textless{}concept1\textgreater cause? | \textless{}concept1\textgreater causes \textless{}concept2\textgreater{}                                         \\ \hline
CausesDesire                    & causes the desire to           & What desire does \textless{}concept1\textgreater cause? | \textless{}concept1\textgreater causes desire of \textless{}concept2\textgreater{}                        \\ \hline
CreatedBy                       & is created by                  & What is \textless{}concept1\textgreater created by? | \textless{}concept1\textgreater is created by \textless{}concept2\textgreater{}                               \\ \hline
Desires                         & desires                        & What does \textless{}concept1\textgreater desire? | \textless{}concept1\textgreater desires \textless{}concept2\textgreater{}                                       \\ \hline
HasPainCharacter                & has pain character of          & What pain character does \textless{}concept1\textgreater have? | \textless{}concept1\textgreater has pain character of \textless{}concept2\textgreater{}            \\ \hline
HasPainIntensity                & has pain intensity of          & What pain intensity does \textless{}concept1\textgreater have? | \textless{}concept1\textgreater has pain intensity of \textless{}concept2\textgreater{}            \\ \hline
InheritsFrom                    & inherits from                  & What does \textless{}concept1\textgreater inherit from? | \textless{}concept1\textgreater inherits from \textless{}concept2\textgreater{}                           \\ \hline
InstanceOf                      & is an instance of              & What is \textless{}concept1\textgreater an instance of? | \textless{}concept1\textgreater is an instance of \textless{}concept2\textgreater{}                       \\ \hline
LocatedNear                     & is located near                & What is \textless{}concept1\textgreater located near? | \textless{}concept1\textgreater is located near \textless{}concept2\textgreater{}                           \\ \hline
LocationOfAction                & has location of action at      & What location of action does \textless{}concept1\textgreater have? | \textless{}concept1\textgreater has location of action of \textless{}concept2\textgreater{}    \\ \hline
ReceivesAction                  & receives action of             & What action does \textless{}concept1\textgreater receive? | \textless{}concept1\textgreater received action of \textless{}concept2\textgreater{}                    \\ \hline
Antonym                         & is an antonym of               & What is an antonym of \textless{}concept1\textgreater{}? | \textless{}concept1\textgreater is an antonym of \textless{}concept2\textgreater{}                       \\ \hline
DerivedFrom                     & is derived from                & What is \textless{}concept1\textgreater derived from? | \textless{}concept1\textgreater is derived from \textless{}concept2\textgreater{}                           \\ \hline
DistinctFrom                    & is distinct form               & What is \textless{}concept1\textgreater distinct form? | \textless{}concept1\textgreater is distinct form \textless{}concept2\textgreater{}                         \\ \hline
EtymologicallyRelatedTo         & is etymologically related to   & What is \textless{}concept1\textgreater etymologically related to? | \textless{}concept1\textgreater is etymologically related to \textless{}concept2\textgreater{} \\ \hline
FormOf                          & is a form of                   & What is \textless{}concept1\textgreater a form of? | \textless{}concept1\textgreater is a form of \textless{}concept2\textgreater{}                                 \\ \hline
HasContext                      & has context of                 & What context does \textless{}concept1\textgreater have? | \textless{}concept1\textgreater has context of \textless{}concept2\textgreater{}                          \\ \hline
SimilarTo                       & is is similar to               & What is \textless{}concept1\textgreater similar to? | \textless{}concept1\textgreater is similar to \textless{}concept2\textgreater{}                               \\ \hline
Synonym                         & is a synonym of                & What is a synonym of \textless{}concept1\textgreater{}? | \textless{}concept1\textgreater is a synonym of \textless{}concept2\textgreater{}                         \\ \hline
dbpediacapital                  & has the capital city           & What is the capital city of \textless{}concept1\textgreater{}? | \textless{}concept1\textgreater has capital city of \textless{}concept2\textgreater{}              \\ \hline
dbpediaproduct                  & has product                    & What product does \textless{}concept1\textgreater have? | \textless{}concept1\textgreater has product of \textless{}concept2\textgreater{}                          \\ \hline
\end{tabular}
}
\caption{Knowledge representation mappings.
}
\label{tab:mappings}
\end{table*}

\begin{table*}[tb]
\centering
\resizebox{\linewidth}{!}{
\begin{tabular}{l|c|c|c|c|c|c|c}
\multicolumn{1}{c|}{Model Variants} & Grammatical  & Coherent   & Engaging & Informative & Specific & Common Sense & \textbf{Avg} \\ \hline
TBS-\textbf{soft}-symbol-NL          & \textbf{53.0/10.0\%} & 46.3/8.7\%          & 48.7/9.3\%       & 41.7/20.6\%          & 51.7/6\%       & \textbf{52/7\%}  & 50.5/10.3\% \\
TBS-hard-\textbf{prompt}-NL              & 50.3/4\%           & 49/7.3\%         & 47/9\%          & 49.4/6\%          & 51/3\%         & 48.3/2.7\%      &   49.2/5.3\%   \\
TBS-hard-symbol-\textbf{QA}          & \textbf{53/6.7\%} & \textbf{53.6/5.6\%} & 51.3/4.7\%       & 51.3/3.7\%          & 51.3/5\%       & \textbf{54/3.7\%}  & \textbf{52.4/4.8\%} \\
TBS-\textbf{soft}-\textbf{prompt}-NL          & 49.3/6.7\%  & 49.78/8.7\% &   51.3/4.7\%    &      50.3/2.7\%    &    49.3/8.7\%   & 48.2/6.7\% & 49.8/5.4\\
TBS-\textbf{soft}-symbol-\textbf{QA}         & 51.5/4.2\%  & \textbf{52.1/3.5\%} &   51.9/4.9\%    &      49.2/6.7\%    &    49.9/2.7\%   & 45.3/6.9\% & 51.8/5.6\\
TBS-hard-\textbf{prompt}-\textbf{QA}         & 48.3/7.2\%  & 49.9/7.7\% &   50/5.2\%    &      49.2/5.7\%    &    48.2/6.6\%   & 47.4/2.9\% & 48.8/6.4\\
TBS-\textbf{soft}-\textbf{prompt}-\textbf{QA}        & 50.1/4.7\%  & 50.2/8.7\% &   49.3/7.9\%    &     48.2/8.7\%    &    48.3/2.7\%   & 49.9/5.7\% & 49.9/7.2 
\end{tabular}
}
\caption{Human evaluation on \textbf{response quality} when comparing different model variants with the base model (hard-symbol-NL).
}
\label{tab:appendix_variants}
\end{table*}	

\begin{table*}[tb]
\centering
\scalebox{0.75}{
\begin{tabular}{ccccccccc}
\hline
\multicolumn{1}{c|}{}                         & \multicolumn{4}{c|}{\textbf{Logical Corruption Average {[}Accuracy/$\Delta$ NLL{]}}}                                                                                                                   & \multicolumn{4}{c}{\textbf{Complete Corruption Average {[}Accuracy/$\Delta$ NLL{]}}}                                                                                              \\ \cline{2-9} 
\multicolumn{1}{c|}{\multirow{-2}{*}{Models}} & DD                                         & ED                                         & MuTual                                     & \multicolumn{1}{c|}{SocialIQA}                                  & DD                                         & ED                                         & MuTual                                     & SocialIQA                                  \\ \hline
\multicolumn{9}{c}{\textit{\textbf{Inference Probing}}}                                                                                                                                                                                                                                                                                                                                                            \\ \hline
\multicolumn{1}{c|}{DialoGPT}           & 0.57/-0.01         & 0.60/0.03 & 0.62/0.03 & \multicolumn{1}{c|}{0.64/0.03} & 0.71/0.15          & 0.77/0.25          & 0.79/0.22 & 0.87/0.40 \\
\multicolumn{1}{c|}{KS-RoBERTa}   & 0.49/-0.00         & 0.50/-0.00         & 0.49/-0.00         & \multicolumn{1}{c|}{0.50/-0.00}         & 0.76/0.23 & 0.79/0.24 & 0.78/0.24          & 0.81/0.27          \\
\multicolumn{1}{c|}{TBS}         & 0.61/0.15          & 0.57/0.07          & 0.57/0.07          & \multicolumn{1}{c|}{0.56/0.05}          & \textbf{0.88/1.38}         & \textbf{0.86/1.24}         & \textbf{0.87/1.14}         & \textbf{0.89/1.47}        \\ \hline
\multicolumn{1}{c|}{Human}                    & 1.0                & 1.0                & 0.9                & \multicolumn{1}{c|}{1.0}                & 1.0                & 1.0                & 1.0                & 1.0                \\ \hline
\end{tabular}
}
\caption{CEDAR~\cite{zhou2021probing} results where bold-faced numbers indicate statistically significant differences comparing to the second-best model.
}
\label{tab:CEDAR}
\end{table*}
\subsection{Models Combining Variants}\label{appendix_variants}
Table~\ref{tab:appendix_variants} presents the complete results considering all of our models' variants. We find that the best overall configuration is hard-symbol-QA.

\subsection{CEDAR Probing: Do TBS models understand why a response makes sense?}
We follow the CEDAR probing framework from~\citet{zhou2021probing} that analyzes if RG models assign a higher probability to the response when provided with valid common sense in the form of explanations compared to corrupted explanations. Results comparing to an end-to-end RG model and a knowledge-selection model are shown in Table~\ref{tab:CEDAR}. We find that by TBS training, RG models become much more sensitive to commonsense explanations against complete corruptions but still fall short against more subtle logical corruptions that require deeper reasoning.

\end{document}